\pdfoutput=1
\documentclass[11pt,a4paper]{article}
\usepackage[hyperref]{eacl2023}
\usepackage{times,graphicx,dialogue,amsfonts}
\usepackage{latexsym}

\usepackage{microtype}

\aclfinalcopy 
\def\aclpaperid{31} 

\title{A Two-Sided Discussion of Preregistration of NLP Research}

\author{Anders Søgaard \quad Daniel Hershcovich \\
Department of Computer Science \\
University of Copenhagen \\
\texttt{\{soegaard,dh\}@di.ku.dk}
\And Miryam de Lhoneux \\
Department of Computer Science \\
KU Leuven \\
\texttt{miryam.delhoneux@kuleuven.be}}

\begin{document}

\maketitle

\begin{abstract}
\Citet{van-miltenburg-etal-2021-preregistering} suggest NLP research should adopt {\em preregistration} to prevent fishing expeditions and to promote publication of negative results. At face value, this is a very reasonable suggestion, seemingly solving many methodological problems with NLP research. We discuss pros and cons---some old, some new: a) Preregistration is challenged by the practice of retrieving hypotheses after the results are known; b) preregistration may bias NLP toward confirmatory research; c) preregistration must allow for reclassification of research as exploratory; d) preregistration may {\em increase} publication bias; e) preregistration may {\em increase} flag-planting; f) preregistration may {\em increase} $p$-hacking; and finally, g) preregistration may make us less risk tolerant. We cast our discussion as a dialogue, presenting both sides of the debate. 
\end{abstract}

\section{Preregistration}

Should NLP researchers be required to preregister their studies? \Citet{van-miltenburg-etal-2021-preregistering} present arguments {\em for}~preregistration, recently echoed by \citet{ulmer2022experimental}. Preregistration has its origin in preregistration of clinical trials,\footnote{The first registries were established by medical researchers in the 1960s and were originally designed to help experimenters recruit participants for clinical trials, but as pointed out by \citet{wiseman2019registered}, preregistration, as we think of it today, started in parapsychology. In 1974, Martin Johnson, a professor of parapsychology and an editor of newly established European Journal of Parapsychology, introduced a preregistration practice for this journal \cite{johnson1975models}, in an effort to make parapsychology protocols more rigorous. In the editorial, Martin Johnson describes how according to the philosophy of the proposed preregistration model, experimenters should define their problems, formulate their hypotheses and outline their experiments, prior to commencing their studies. In Declaration of Helsinki \S19, the \citet{association2013world} demands: ``Each clinical study must be registered in a publicly accessible database before the first test subject is recruited.'' While the European Commission refers to it, it has not been universally adopted \cite{rid_schmidt_2010}.} and amounts to the following: Before you initiate a set of experiments, you register your hypotheses, your experimental design and how you plan to analyze your results. Registration is time-stamped on an online platform with general public access. You then follow your plan as closely as possible and report any divergences in your final publication. 

The discussion in \citet{van-miltenburg-etal-2021-preregistering} is not unprecedented. Preregistration has been debated in epidemiology \cite{lash2012should}, social psychology \cite{veer2016preregistration}, experimental economics \cite{STROMLAND2019102143} and information systems research \cite{bogert2021preregistration}. Our discussion is inspired by the discussion in epidemiology, which is similar to NLP in focusing on data analysis rather than clinical trials.

There is an important ambiguity in how preregistration is discussed: Is the preregistration entry peer-reviewed or not? \citet{chambers2019whatsnext} sees preregistration as a peer-reviewed process, and this is also what \citet{van-miltenburg-etal-2021-preregistering} suggest for NLP. We therefore assume peer-reviewed preregistration below. The required format of the registered report is also important. In their Appendix, \citet{van-miltenburg-etal-2021-preregistering} provide example questionnaires. We will assume registered reports will be lists of answers to such questionnaires, but in \S9, we will suggest a few revisions to the questions formulated by  \citet{van-miltenburg-etal-2021-preregistering}. 

\section{Why Preregister NLP Research?}\label{sec:why}

\Citet{van-miltenburg-etal-2021-preregistering} present four reasons for adopting preregistration in NLP: distinguishing between confirmatory and exploratory research, avoiding fishing expeditions and harking, mitigating publication bias and avoiding flag-planting: 

\paragraph{Distinguishing Confirmatory from Exploratory} The first apparent advantage to preregistration---often said to be the most important one \cite{Nosek2600}---is that it clarifies what counts as confirmatory research, which {\em has to} preregister, and what counts as exploratory research with no obligation to preregister. Confirmatory research is hypothesis testing, held to the highest standard and which aims to minimize false positives. Here, $p$-values are generally assumed to have diagnostic value and inferences can be drawn to wider populations. Exploratory research, in contrast, has a different status: It generates rather than tests hypotheses and results should be replicated and confirmed at a later stage. Typically, the focus is on minimizing false negatives, and $p$-values are not assumed to have diagnostic value \cite{https://doi.org/10.1111/1740-9713.01369}. Moreover, findings are not assumed to be directly transferable to wider populations.  \citet{TQMP16-4-376}, however, points out how it is not always trivial to distinguish between confirmatory and exploratory research: if a researcher, for example, retries a hypothesis from previously published literature to explain an experiment they just ran, is this an {\em a priori} or a {\em post-hoc}~hypothesis? See also \S3. 

\paragraph{Fishing Expeditions and Harking} Preregistration is often said to prevent {\em fishing expeditions} and so-called {\em harking}\footnote{Short-hand for ``hypothesizing after the results are known''. Often conflated with fishing, but the two differ: Harking fixes the experiment, varies the hypothesis, so to speak, whereas fishing fixes the hypothesis, varies the experiment.
 The acronym was coined by social psychologist Norbert Kerr.
} \cite{chittaranjan21hark}, namely, post-hoc characterization of hypotheses based on experimental outcomes. Fishing expeditions is ambiguous in the literature (between fishing and harking), but we use the term to refer to cherry-picking dataset and protocols to validate a hypothesis. Harking, in turn, is what researchers do when they indiscriminately examine associations between different variables, not with the intention of testing {\em a priori} hypotheses but simply hoping to find something of significance. \citet{TQMP16-4-376} calls this `undisclosed hypothesizing after the results are known.' Having authors preregister their hypotheses potentially improves the reliability of confirmatory research by controlling for cherry-picking and multiple hypothesis testing, implicit to exploratory research. See also \S4. 

\paragraph{Publication Bias} \Citet{van-miltenburg-etal-2021-preregistering} say that, to them, the main advantage of registered reports is that they
provide a means to avoid publication bias. Because studies are evaluated {\em prior}~to the results, negative results have the same chance to be published as positive ones. \citet{TQMP16-4-376} refer to this as avoiding {the suppression of {\em a priori} hypotheses that yield null or disconfirming results}. Publication bias is claimed to be a serious problem in NLP research by many \cite{plank-etal-2014-importance,card-etal-2020-little,cohen-etal-2021-reviewing}. The argument was also used in epidemiology, but received some pushback \cite{loder10registration}. See also \S5. 

\paragraph{Flag-planting} \Citet{van-miltenburg-etal-2021-preregistering} also suggest preregistration can prevent so-called {\em flag-planting}. Flag-planting refers to rushing to be the first to publish results. Flag-planting potentially comes at the cost of scientific integrity and quality. Because of biases in peer-reviewing, it is harder to publish a corrected version of a study that is already out there, than to publish an error-prone study that is the first of its kind. See also \S6. 

\paragraph{Other Reasons to Preregister} We have covered the main reasons \citet{van-miltenburg-etal-2021-preregistering} had for adopting preregistration and will now move to our two-sided, dialogical discussion of its pros and cons. In our dialogue, we will let Zeny and Socart,\footnote{Zeny is a mix of Kenny from South Park and Zeno. In Plato's {\em Parmenides}, Zeno argues for monism---the idea that reality is one stable thing. Socart is a mix of South Park's Eric Cartman and Socrates, who countered this idea by asserting a more nuanced ontology in which things stand in complex relations to each other. Socrates, in other words, took a more nuanced stance, arguing against the existence of a one-size-fits-all hypothesis. Zeny and Socart adopt similar positions in our dialogue about preregistration.} our house philosophers, debate preregistration. In \S3--\S8, we will let them discuss arguments {\em against} preregistration, including arguments that run counter to those presented by \citet{van-miltenburg-etal-2021-preregistering}, but we will first let Zeny provide us with a fifth argument {\em for}~preregistration:

\begin{small}\begin{dialogue}

\speak{Zeny} Socart, there's an additional argument for preregistration, I believe. Early feedback on experimental methodology through a peer-reviewed registration process should improve the quality of the methodology, should it not? Such feedback also saves resources otherwise spent on failed or misleading experiments.

\speak{Socart} Zeny, we both know turn-around is fast in NLP research. Experiments are easier to run and feedback is much faster than for clinical trials, where preregistration is common.

\speak{Zeny} NLP as a field has many virtues, but the reviewing cycle is slowing as the field grows larger. Moreover, experiments are becoming more expensive with larger models, creating barriers of entry \cite{10.1145/3442188.3445922} and experiments have substantial environmental impact.

\speak{Socart} You make an important point, Zeny, but early feedback would require more time from reviewers. Since reviewers and researchers coincide, preregistration would potentially save compute resources, but not working hours.  

\speak{Zeny} That is an oversimplification. Giving feedback on an early draft takes much less time than writing a full paper. If the reviewers are carried over, they will save time when reading the full submission, also. Preregistration would also prevent cherry-picking and invalid use of significance tests by excluding explorations from confirmatory research.

\speak{Socart} But the explorations could be done prior to preregistration and researchers may then be more inclined not to report such explorations at all.\footnote{This point was also made for preregistration in epidemiology by \citet{ae2168802dc011df9806000ea68e967b}.}

\speak{Zeny} Any system can be tricked, but if researchers adopted the practice of preregistration, we would, all things being equal, increase transparency and decrease bias around research.

\speak{Socart} Dear Zeny, you too have seen the evasiveness of bureaucracy, e.g., in NLP conference submission forms. While preregistration reports would initially be light-weight, transparency could easily be clouded by the complexity of assembling the information required for preregistration as new requirements are added over time.\footnote{See \citet{loder10registration} for arguments from epidemiology.} 
\end{dialogue}\end{small}

\noindent See \citet{bracken2011} and \citet{TQMP16-4-376} for a discussion of more advantages to preregistration.
In addition to reducing fishing expeditions and harking, flag-planting and publication bias, these include: {\bf a)} preventing $p$-hacking, {\bf b)} prespecifying tolerated significance levels, {\bf c)} identifying selective reporting,\footnote{Selective reporting is regarded the most
important contributor to irreproducibility by \citet{baker2016isthere}. \citet{Nosek2600}, advocating for preregistration, presents similar arguments.} and {\bf d)} preventing {\em forking paths} practice.\footnote{This practice refers to when researchers make decisions about which correlation tests to conduct based on properties of their data. The practice is named after {\em The Garden of Forking Paths} 
, a 1941 short story by Jorge Luis Borges. 
} None of these points are uncontroversial and \citet{TQMP16-4-376} also presents counter-arguments against {\bf a)-d)}. For example, prespecified significance levels have been superseded by the practice of simply reporting {\em actual}~$\alpha$-levels.  Surprisingly, there has been little work on whether preregistration increases trust in science, except for the study by \citet{a88772405f594d588336d281cfdf7908}, which 
was under-powered. 

We focus on {\em preregistration for NLP research}. In general, there is no {\em a priori}~reason to think that the pros and cons of preregistration transfer from clinical trials over epidemiology to NLP research. In clinical trials, for example, it is easy to decide when a protocol must be registered. This simply happens before the first subject is assigned to treatment. In epidemiology, there is no such bright line \cite{lash2012should} and it is equally hard to see one in NLP. While general machine learning has seen many related methodological discussions \cite{https://doi.org/10.48550/arxiv.1904.07633,https://doi.org/10.48550/arxiv.1807.03341,gundersen2022sources}, there has, to the best of our knowledge, been no published discussions of preregistration practice in this field, with the exception of \citet{Gundersen_2021}.\footnote{Workshops on preregistration at ICCV 2019 (\url{https://preregister.vision/}) and at NeurIPS 2021 (\url{https://preregister.science/}) seemingly did not lead to publications or a change in practice yet, but the website for the 2021 workshop says papers are forthcoming.}\footnote{\citet{Gundersen_2021} complains no AI venues support preregistration, but provide no arguments for or against it.}

In our discussion below, we will ignore the most trivial challenges to preregistration, such as deviations from data collection plans for practical reasons, discovery of assumption violations, etc. Such challenges have already been discussed in the clinical literature, e.g., by \citet{Nosek2600}.

\section{Encouraging Confirmatory Research}\label{sec:confirmatory}

We let {\sc Socart} and {\sc Zeny} discuss whether preregistration will succeed in distinguishing between confirmatory and exploratory research. A decade ago, when preregistration was being implemented and discussed in epidemiology, the worry that preregistration would introduce a bias against ``the end of the research spectrum that constitutes the quirky, brilliant work that is not enterprise-driven'' \cite{ae2168802dc011df9806000ea68e967b}, was the main concern among its opponents. {\sc Socart} and {\sc Zeny} discuss the consequences of insisting on a distinction that is not trivial to uphold in practice. 

\begin{small}\begin{dialogue}

\speak{Socart} It seems to me, dear Zeny, that many NLP projects are driven {\em not} by an explicit hypothesis, but by a desire to understand the behavior of a model, to be able to characterize its strengths and weaknesses, or by a simple gut feeling that at the locus of interacting variables, interesting dynamics can be observed. 

\speak{Zeny} Can you provide me with an example?

\speak{Socart} Certainly. \citet{pires-etal-2019-multilingual}, e.g., showed that knowledge encoded in multilingual BERT \cite{devlin-etal-2019-bert}, could be transferred across languages---even across scripts, that such transfer worked best between typologically similar languages, that it could process code-switching and find translation pairs. They also showed systematic deficiencies affecting some language pairs. How would they have foreseen these findings? Or even the dimensions that turned out to be of interest? Even if they had foreseen how they wanted to explore transfer across scripts and typological classes, what if genealogy or demography turned out to be more interesting than typology?\footnote{Independent language families may share features, i.e., be typologically close, but genealogically apart. See \citet{rama-kolachina-2012-good} for discussion.}

\speak{Zeny} I am unconvinced that preregistration would be a serious obstacle to such work.  \citet{pires-etal-2019-multilingual} could have defined the search space in advance -- or maybe this is exploratory work that would not have to register in the first place? Remember also, Socart, that the preregistered plan can be updated and refined in the course of a research project. Plans can be revised, but this does not cancel out the benefits of planning.

\speak{Socart} Preregistration may accommodate deviation from the plan, but would risk losing its benefit if researchers were allowed to preregister too many hypotheses or update their plans too frequently. Let us illustrate this with another example. \citet{zhao-bethard-2020-berts} study how BERT models' learned self-attention functions change during fine-tuning to reflect the target task. They find this to be the case only in smaller models; with more parameters, the change disappears. Imagine now that their hypothesis was confirmed only for select combinations of positional encodings, regularizers and optimizers.

\speak{Zeny} This sounds suspiciously like a case of forking paths. \citet{dror-etal-2017-replicability} warned us about this risk, encouraging us to at least validate our hypotheses on multiple datasets to reduce the chance of $p$-hacking.\footnote{See \citet{belz-etal-2021-systematic} for a similar discussion.} Again, preregistration would not be required for all research. 

\speak{Socart} So if authors submitted exploratory work for peer review, would reviewers then decide if bypassing preregistration was appropriate?

\speak{Zeny} Yes. Preregistration clarifies the distinction between exploratory and confirmatory research. 

\speak{Socart} But what if \citet{pires-etal-2019-multilingual} had pointed to earlier work already hypothesizing that transfer works best between typologically similar languages? Would this not have made their research confirmatory in the eyes of their readers and therefore in need of preregistration?\footnote{This practice is known as ``retrieving hypotheses after the results are known'' (r-harking) \cite{rubin2017when}.} 

\speak{Zeny} It very well might have. If they consider it exploratory, they should also point to alternative hypotheses that would explain different results.

\speak{Socart} Moreover, if preregistration becomes a badge of honor or increases your chances of getting your work accepted, because the findings have a different air of trustworthiness,\footnote{Greater reliance on preregistration improves estimation of effect sizes, as shown by \citet{STROMLAND2019102143}.} would this not be a reason to encourage your students to perform confirmatory rather than exploratory research? Preregistration would, in other words, inject a bias toward confirmatory research into NLP.

\speak{Zeny} I, for one, would welcome this kind of bias.
\end{dialogue}\end{small}

\noindent$\checkmark$ {\em Preregistration is challenged by r-harking and may bias NLP toward confirmatory research.}

\section{Some Expeditions May Prevent Others}\label{sec:fishing}

If you ask an NLP researcher if they are ``on a fishing expedition'' or if they are hypothesizing after the fact, you will instantly make them feel very uncomfortable. It is widely accepted that fishing and harking are bad practices.\footnote{\citet{chittaranjan21hark} notes that fishing expeditions can be ``ethical'' if acknowledged as such, and if appropriate corrections are performed when computing significance results.} Socart, however, has an argument {\em for} (occasional) harking: 

\begin{small}\begin{dialogue}

\speak{Socart} Two researchers, Ann and Bob, have the same hunch, that the regularization technique $R_2$ is better than its competitors, $R_0,R_1,R_3$. Ann realizes after a set of experiments of datasets $D_0,D_1,D_2$ that, in fact, $R_1$ is better than $R_2$. Since this was previously unknown to the community, she publishes it, presenting it (somewhat vaguely) as a confirmation of an {\em a priori}~hypothesis. Bob tests the same hypothesis, i.e., that $R_2$ is superior to $R_0,R_1,R_3$. Seeing $R_1$ is better on $D_0,D_1,D_2$, he looks for more datasets, until he has a suite of datasets $D_4,D_5,D_6$ on which $R_2$ is better than $R_1$.  
\speak{Zeny} Bob's cherry-picking is extremely problematic, but so is Ann's harking. 
\speak{Socart} But would you agree that granting her the freedom to hark most likely reduces her temptation to cherry-pick?
\speak{Zeny} This would simply reclassify Ann's work as exploratory rather than confirmatory. I see no reason why preregistration should not allow this. 
\speak{Socart} The two researchers both departed from their original plans, but Ann's willingness to depart from her original hypothesis serves us better than Bob's cherry-picking. In this way, harking can prevent a researcher from taking on a fishing expedition.
\end{dialogue}\end{small}

\noindent$\checkmark$ {\em Preregistration should allow for re-classification of confirmatory research as exploratory research.}\footnote{Reclassification flags work as exploratory, thereby increasing transparency, but would not impact acceptance decisions. An alternative, suggested by one of our reviewers, would be to introduce intermediate reports as a required step to share the results of the preregistered study before continuing to preregister and test alternative hypotheses a part of the same study. This further increases transparency, and prevents having `unwanted' results `swept under the rug' in the final publication. Researchers working on a similar topic would already benefit from the results in an intermediate report.}

\section{Solving Publication Bias?}\label{sec:pubbias}

\citet{ae2168802dc011df9806000ea68e967b} argue against preregistration solving publication bias, because researchers can still selectively register studies after preliminary data explorations. Imagine Hippocrates, the Greek physician, was asked to preregister his vivisection experiments. If Hippocrates was studying 10 soldiers with brain lesions, what would prevent him from using one soldier to generate hypotheses, preregister those and conduct the final experiments on the remaining nine? Or worse, peek at all, preregister and go back to the data?

\begin{small}\begin{dialogue}

\speak{Socart} Say Hippocrates has two hypotheses about the soldiers, such as that the heart is the seat of compassion and that the brain is the seat of rational thought. Upon preliminary exploration, he sees many soldiers have turned cold-hearted by the atrocities of war, but few complain of heartaches. Nearly all soldiers who are delusional or suffer from memory loss, also suffered blows to their heads. Hippocrates pursues and preregisters only the hypothesis that the brain is the seat of rational thought. He has now preregistered, not a prediction, but a post-diction, ignoring the negative result. 

\speak{Zeny} But Socart, did you not, a moment ago, argue that preregistration would dampen the creativity of research by preventing fishing expeditions and harking? 
\speak{Socart} In theory, yes. Preregistration will dampen creativity if properly sanctioned, but I am skeptical that this would be practically possible, rendering preregistration ineffective, an unnecessary administrative burden for all---and a bottleneck for the honest few. 

\speak{Zeny} If preregistration {\em prior} to data collection is encouraged, this would solve the problem, no?\footnote{This is implicit in preregistration in scientific fields where data is not re-used, e.g., in psychology \cite{wiseman2019registered}.}

\speak{Socart} Surely, but this would mean only one preregistered study per dataset. Since few NLP papers introduce new datasets, this would render preregistration ineffective for the vast majority of NLP research. 

\speak{Zeny} This is a good point, Socart, but community-wide overfitting to benchmarks is a vice, not a virtue. If preregistration encourages the introduction of new test datasets, that's a good thing, no? Some even argue that all papers should ideally introduce new test data. 
\end{dialogue}\end{small}

\paragraph{To the contrary?}\ldots in which {\sc Socart} and {\sc Zeny} continue to  discuss whether preregistration could actually make publication bias worse.  {\sc Socart} suggests that preregistration could amplify publication bias, if positive results are still preferred over negative ones and preregistration forces researchers to focus on predictably positive results, arguably a small subset of the positive results. If papers are accepted on the basis of preregistration, this could increase an arguably already existing bias toward incremental improvements.\footnote{\citet{lash2012should}, for example, argue that: ``prespecified hypotheses often take little risk, invoke little imagination and stray only a short distance from what is already well understood.''} 

\begin{small}\begin{dialogue}

\speak{Socart} You say preregistration will make it easier to publish negative results, because studies are evaluated prior to obtaining results? 

\speak{Zeny} That is correct, Socart. 

\speak{Socart} \ldots but do we really know why there are so few NLP papers about negative results? See, in NLP, negative results are much harder to establish than positives. If I want to show that self-attention or weight averaging does {\em not} lead to improvements for some problem, I need to show that holds across all implementations, all architectures and all available datasets. The value of a report stating that for one such combination, self-attention didn't do much, would be next to nothing. Isn't that the real explanation for the skew in the NLP literature?

\speak{Zeny} Negative results are key to scientific progress \citep{10.3389/fnins.2019.01121}, but are hard to establish if they are very general. Published positive results often overclaim their generality.  Both should be confirmed only by accumulated evidence in diverse settings.\footnote{NLP has seen relatively few meta-studies \cite{cramer-2008-well,sogaard-2013-estimating,DBLP:conf/nips/HoyleGHPBR21,bugliarello-etal-2021-multimodal}, but hopefully, we will see more in the future.}

\speak{Socart} No, no, you fail to see there's a qualitative difference, Zeny! Imagine if Ann was evaluating self-attention for sentiment analysis. To answer the hypothesis that self-attention works in the positive, she just needs a significant result in a single setting. In contrast, in order to establish a negative result, she has to explore {\em all} possible settings.\footnote{We flesh this out a bit. The research hypothesis in Ann's case is that self-attention helps. What this means is that in {\em some} implementation, it leads to robust improvements. The vast majority of NLP hypotheses take this form: $X$ can, in some implementation, lead to general improvements on one or more tasks. If the baseline is fixed, this amounts to existential quantification (``some''). Conversely, its negation (``self-attention does not help for sentiment analysis'') amounts to universal quantification, i.e., there is {\em no}~implementation in which this is the case. This is obviously much harder to prove than the corresponding positive result.} How would preregistration make establishing a negative result less formidable a challenge? 

\speak{Zeny} I agree that it would not. My only claim is that evaluating studies prior to obtaining results would prevent any bias on behalf of peer reviewers to evaluate negative results more harshly.

\speak{Socart} \ldots but in reality, we do not know if such a bias exists, or whether it is only fair that such a bias exists, because the bar by definition should be higher for negative results?
\end{dialogue}\end{small}

\noindent$\checkmark$ {\em Preregistration may increase publication bias.}

\section{Solving Flag-Planting?}\label{sec:flags} 

Flag-planting is one of the motivations for preregistration for \citet{van-miltenburg-etal-2021-preregistering}, but \textit{exclusive} preregistration may also, conceivably, have the opposite effect. 

\begin{small}\begin{dialogue}

\speak{Socart} Say Ann and Bob get the same great idea---e.g., to evaluate the sensitivity of textual entailment models to presupposition projection---and worry that they will be scooped before getting around to publishing it. Ann and Bob now follow two different strategies: Bob rushes to preregister a study hypothesizing that state-of-the-art models are sensitive to such phenomena, while Ann rushes to run the experiments and publish the paper. Zeny, which strategy is better for science?

\speak{Zeny} I would say it's Bob's, since rushed experiments are more likely to be flawed.

\speak{Socart} But Bob plants his flag faster than Ann, essentially scooping her. By doing so, Bob discourages Ann from pursuing this idea by planting his flag first. What if Bob fails to conduct proper experiments altogether? Had it not been for preregistration, both researchers would have pursued their idea, providing mutual replication and increasing the likelihood of the idea materializing into an actual result.

\speak{Zeny} Some would say this is one of the advantages of preregistration: Ann pursuing the same idea would have been a waste of time. 

\speak{Socart} This assumes Ann and Bob would have conducted their research in exactly the same way and that none of them were prone to error. In other words, that researchers are machines that simply execute their unambiguous experimental protocols. I think preregistration just moves flag-planting to earlier in the research process, lowering the bar for researchers to plant their flags, since less work is required to plant a flag. And when a bar is lowered, more researchers are likely to plant more flags. 

\speak{Zeny} \Citet{van-miltenburg-etal-2021-preregistering} explicitly encourage concurrent work.

\speak{Socart} Yes, I did read that passage, but they do not discuss {\em how} preregistration would impact concurrent work. Do they envisage a review system in which Ann is allowed to follow up on the idea that Bob preregistered?

\speak{Zeny} I'd do that, in the spirit of open science.

\speak{Socart} Such \textit{inclusive} preregistration would clearly discourage protectionist researchers from preregistering their studies. If a preregistered study is up for grabs for other research labs, labs with more resources could likely wrap up the experiments faster than the researchers who registered it.

\speak{Zeny} \ldots unless we envisage a review system allowing Ann to preregister the same study, giving Ann and Bob equal chances to pursue the study.

\speak{Socart} This would be equivalent to telling reviewers of a paper to consider as ``concurrent'' any other work published within the last 1--2 years (assuming this is the approximate life span of a research project), including preregistered studies. Today, reviewers are told to disregard work published within the last three months, but already, reviewers seem to ignore this guideline in practice, presumably because they do not want to compromise the fast turn-around in NLP research.

\speak{Zeny} But shouldn't we incentivize slow science? Many NLP papers neglect related work and keep reinventing the wheel. We need deeper analysis to enable disruptive scholarship and novel ideas.\footnote{\citet{doi:10.1073/pnas.2021636118} showed that fast turn-around results in stymied fundamental progress in large scientific fields.}

\speak{Socart} Slow science also has disadvantages. Fast turn-around has had many positive effects on NLP, including rapid replication. Projects can become ``too big to fail,'' causing confirmation bias.\footnote{This can result from financial interests \citep{10.1371/journal.pmed.0020124}, e.g., due to ``sunk cost'' \citep{https://doi.org/10.1002/anie.202208429}.} Lowering false positive rates is important, but so is healthy distrust in published results. 
\end{dialogue}\end{small}

\noindent$\checkmark$ {\em Preregistration may increase flag-planting.}\footnote{One obvious solution is to make preregistration non-public, but then preregistration would not prevent two groups doing the same study.} 

\section{Solving $p$-Hacking?}\label{sec:phack}

Inflation bias, also known as $p$-hacking, refers to selective reporting to produce statistically significant results. \citet{sogaard-etal-2014-whats} lists several $p$-hacking techniques used, perhaps inadvertently, in NLP papers. If a statistically significant result is seen as the key to getting your paper accepted, researchers are presumably willing to go far to squeeze out a small $p$-value. But if preregistration facilitates the publication of negative results, it seems it would also reduce the incentive to engage in so-called $p$-hacking, e.g., obsessive fiddling with data and models until reaching the magical $p<0.01$. It has been noted, however, that preregistration leaves plenty of room for $p$-hacking \cite{bakker2020}. Generally, eliminating $p$-hacking entirely is unlikely when career advancement is assessed by publication output, and positive results are favored by scientific peers \cite{10.1371/journal.pbio.1002106}.

Socart and Zeny discuss whether preregistration will reduce or amplify the incentive to engage in $p$-hacking: 

\begin{small}\begin{dialogue}

\speak{Socart} Imagine Ann again, who is now evaluating if self-attention is helpful for sentiment analysis. Say she preregisters the hypothesis that self-attention {\em is} helpful, only to find that her first results are negative. We would now like Ann to go ahead and acknowledge the negative results on print, right? However, as we just saw, when your first results are negative, more results are typically needed to draw a firm, negative conclusion that self-attention does {\em not} help. Sometimes more data collection is needed and more human evaluations may be needed. Pursuing the negative result will, in other words, be a lot of work. 

\speak{Zeny} But very important!

\speak{Socart} Preregistration increases the amount of work that goes into moving your focus to establishing a negative result: You will need to augment your preregistration with information about the experiment, your new hypothesis and the new experiment you plan to perform. 

\speak{Zeny} Documentation has to be light-weight. 

\speak{Socart} \ldots but preregistration would get people more invested in their ideas and bias them in how results are interpreted. When people go on record with a study description, they will defend why it's reasonable and likely leading to a positive result. Researchers are always prone to confirmation biases, but now social expectations and reputation will amplify their existing biases. This would lead to the opposite of the effect intended. 
\end{dialogue}\end{small}

\noindent$\checkmark$ {\em Preregistration may increase $p$-hacking.}

\section{Risk Tolerance} 

Attempts to reduce false positives tend to also lead to reductions in true positives. Many applications require near-zero false positive rates, but most NLP experiments show low risk of direct negative impact on society or individuals therein,\footnote{This does not refer to the downstream risks after deployment, just the risks associated with the research experiments. Two reasons for NLP experiments being relatively low risk are the rare involvement of human participants in NLP experiments and the historical focus on professionally generated text \cite{hovy-spruit-2016-social}. We are seeing a shift toward human-in-the-loop evaluations and user-generated content, but this still makes for a small fraction of NLP research.} as indicated by the relatively few papers receiving ethical reviews. Hence, we can afford to take risks and explore hypotheses that end up wrong. \citet{parascandola2010epistemic} reminds us how this is a key ingredient in increasing knowledge and reducing uncertainty, getting us off the beaten track. NLP benefits from being frequently wrong and implementing preregistration to prevent false positives has a drawback. In \S9, we will argue that what is important is to balance preregistration with our risk tolerance. 

\begin{small}\begin{dialogue}

\speak{Socart} Imagine Ann works on hate speech detection for a social media company. Bob works on topic classification of social media posts at the same company. They both validate and evaluate the models in the wild on beta users. They both can use logistic regression and SVMs. SVM is sometimes superior, but exhibits more variance across hyper-parameters. If I were to advise Ann or Bob to use logistic regression, who then?

\speak{Zeny} Probably Ann, since we can tolerate less risk in her situation. But how would preregistration affect the risk tolerance of researchers? 

\speak{Socart} Imagine if you were asked today to carry a solid bottle of olive oil over to Plato's house and tomorrow to bring him a fragile, beautiful vase decorated with gold. On which of the two days would you be more inclined to {\em run} there?

\speak{Zeny} Today, but how is this relevant? What are the tasks where we can afford to `run'? 

\speak{Socart} For tasks in which false positives are associated with high risk, we should hedge our bets by preregistering conservative hypotheses; for other tasks, this sort of inhibition of  is unfortunate. 

\speak{Zeny} This is exactly why exploratory research still has a place in a world with preregistration---namely, for tasks where we can tolerate risk.
\end{dialogue}\end{small}

\noindent One may argue that risk mitigation is not what preregistration is for. The purpose of institutional review boards (IRBs) and ethics reviewing is to flag and prevent too risky studies (IRBs focus on risk to human participants, while ethics reviewing also addresses potential applications). We have three reasons to think preregistration requirements should depend on expected risk: (a) It is impossible to review the implications of a study before you have a solid study plan. If preregistration includes risk assessment, this could provide input for IRBs and ethics reviewing (or, in a more distant future, be part of the same process). (b) A partial roll-out of preregistration may help us balance Type 1 and Type 2 errors. Expected risk affects the cost of false positives and hence the optimal balance between Type 1 and Type 2 errors. Since bureaucracy, by the end of the day, also incurs a cost on society, this reinforces our belief that mechanisms should be implemented only for where there is direct impact on society at large. (c) Finally, ethics reviewing is typically part of the standard review process, i.e. {\em after the fact} and can therefore not respond to malpractice in the experimental design or prevent publication of preprints. 

Overly cautious preregistration practice may, in sum, decrease our true positive rate and add bureaucratic overhead to research practices without proper motivation. An all-over-the-map roll-out of preregistration would change the risk tolerance in research and society, just like registration and documentation has increased risk sensitivity in the past. Simultaneously, evaluating risk early on has clear advantages over the current review process.\\

\noindent$\checkmark$ {\em Preregistration may lower our risk tolerance.} 

\section{A Proposal}

We have tried to present pros and cons of preregistration. If we have focused a bit more on the cons, this is only because \citet{van-miltenburg-etal-2021-preregistering} did a great job highlighting its advantages. We will, if anything, argue for only a partial roll-out of preregistration of NLP research. Preregistration is a way to minimize harms of NLP research, but only when risk is high. To motivate this, consider, as first noted by \citet{lash2012should}, two seemingly opposed arguments {\em for}~preregistration: a) Preregistration counters the suppression of (negative) results. b) Preregistration identifies false positives. \citet{lash2012should} argue that while (b) is a valid argument for preregistration of {\em clinical trials}, it is not a valid argument for preregistration in the context of mere ``accumulation of evidence'' \cite{lash2012should}. Here, concerns about balancing Type 1 and Type 2 errors disappear. Preregistration mitigates risks associated with research, reducing potential harms, but at the cost of scientific progress. This calls for a cost-benefit analysis: How much risk can be tolerated for what potential gains?

One way to frame this discussion in NLP is to ask how afraid we should be of being wrong. In clinical trials, there is a significant cost to being wrong. In biomedical studies, false positive rates have been found to be around 14\% \cite{Jager2014AnEO}. Whatever the number is for NLP, lowering it by adding more checks, will lead to a drop in the {\em true} positive rate. If a false positive could result in human tragedy, the price of a lower true positive rate is worth paying, but in NLP, the cost of a false positive is often paid for in compute and human hours. While both can be scarce resources, the open access nature of NLP makes being wrong less dangerous, since mistakes are quickly corrected.\footnote{Of course researchers are sometimes blind-sighted by scientific paradigms and hype, biasing their interpretation of results. Such dynamics is central to, e.g., the Popper-Kuhn debate \cite{rowbottom2011kuhn}, but beyond the scope of this paper.} Zeny would object that pretraining of language models is not easily reproducible \cite{10.1145/3442188.3445922}. Pretraining very large language models should maybe be required to preregister and this would possibly require revising the questionnaires provided by \citet{van-miltenburg-etal-2021-preregistering}. Another concern is how NLP contributes to social and cultural inequality \cite{hershcovich-etal-2022-challenges}. If NLP research is likely to help some more than others, this may be reason to require preregistration. Here, the questionnaires provided by \citet{van-miltenburg-etal-2021-preregistering} would also be insufficient.\footnote{Specifically, we think the questionnaire for 'NLP Engineering experiment paper' (\S{A.3}) should include questions about computational resources needed for pretraining. Since the risk of wasting resources is high for language model retraining, preregistration and early feedback may be particularly useful for such research, but the review of the registered point would have to take this information into account. To mitigate social and cultural inequality, we propose to revise the questionnaire for 'Resource paper' (\S{A.6}), adding questions about the demographics of data sources and annotators, as well as making a corresponding explication of social and cultural concerns in Question 11 of \S{A.3}.} 

So what we propose here reflects a middle-of-the-road position on preregistration. The idea is to {\bf limit preregistration to research for which our risk tolerance is low}. This prevents most of the adverse effects of preregistration, e.g., publication bias, flag-planting and $p$-hacking. NLP research is subject to IRB and ethics reviewing, but we believe this should be merged with preregistration (\S8). Reports should be reviewed and reviewers follow the submission (\S2).\footnote{This would be hard to coordinate for most fields, but in NLP, the ACL Rolling Review platform could make it easier.} Currently, we accept and reject papers through blind peer-reviewing, but some papers are accepted {\em conditional} on a positive ethics review. We propose a reviewing procedure in which some work is only accepted conditional on the work having already been registered with positive reviews. For researchers, this would mean you need to get your preregistered reports accepted, before you initiate the research project. Once completed you will send the final submission in for a new set of reviews, hopefully by the same reviewers. This procedure is somewhat cumbersome and has all the disadvantages we discussed above. Therefore, it should only be used when it is deemed necessary, i.e., when the expected risk of the NLP research is sufficiently high. 

A paper which a) is confirmatory and b) concerns an application for which risk tolerance is low, can be rejected for not having preregistered. We noted the necessity of allowing preregistered research to re-classify as exploratory, i.e., {conditional acceptance for non-preregistered, flagged research, if it explicitly labels itself as exploratory}. This would lead to three different categories of accepted papers: (i) non-preregistered (confirmatory or exploratory) research, (ii) preregistered, confirmatory research and (iii) non-preregistered research {\em marked explicitly as exploratory}. This would, we argue, give us the advantages of preregistration where they are most needed, e.g., where false positives are associated with very high risk. 

We left one important thing in the open for now: How do we fairly decide if a research subject and protocol warrants low risk-tolerance? Ethics review board members are already asked to flag work that `exhibits an increased risk of harm outside the current norms of NLP or CL research'.\footnote{\url{https://aclrollingreview.org/ethicsreviewertutorial}} 
This can be hard to determine, but board members {\em already} have to make this difficult decision. 
Ethics reviews could learn from established risk assessment frameworks \cite{https://doi.org/10.48550/arxiv.2011.04328}.

\section{Conclusion}

Our two-sided dialogue has discussed pros and cons of preregistration in NLP, building on similar discussions in epidemiology. What opponents elsewhere have proposed as alternatives to preregistration is already found in NLP research: open access, common repositories and data sheets \cite{lash2012should}. Preregistration, we argue, is less urgently needed in fields that already facilitate {\bf replication} and where {\bf risk of false positives} is  low. The {\bf fast turn-around} of NLP research means the advantages of transparency and early feedback are smaller. Nevertheless, society's risk-tolerance varies across NLP applications. Legal or medical decision support systems are high-risk application areas. Here, we need to consider all safety measures on the table, including preregistration. 

\section*{Impact Statement}

Preregistration is one of several practices that promote responsible, high-quality research. Others include replication, transparency and open access, as well as impact statements and explicit discussion of study limitations. All such practices come with pros and cons and it is key to scientific progress and positive impact that scientific communities evaluate which practices are adequate in their domain.
The increasing real-world impact that NLP research has exhibited recently and will likely continue to exhibit warrants a careful reconsideration of which practices are called for. Since a major driver of the same impact is the fast-paced exploratory research that characterizes the field, limiting such research may have negative effects as well (see \S\ref{sec:flags}). We therefore believe our two-sided debate will enable an overall better outcome in terms of impact.

\section*{Limitations}

Our discussion of preregistration is inspired by discussions in epidemiology. Many of the concerns epidemiologists had with preregistration seem more relevant to NLP research than the considerations that, by and large, led clinical research to adopt preregistration as a mandatory practice. While we present a proposal for how to implement preregistration in NLP in \S9--a proposal that differs from the one presented by \citet{van-miltenburg-etal-2021-preregistering}--our main contribution is a two-sided discussion of its pros and cons, leaving many questions in the air. Our paper is intended to get the preregistration debate off ground, not to nail it to the floor. 

\section*{Acknowledgements}

Thanks to Asbj{\o}rn Hr\'{o}bjartsson and Klemens Kappel for providing us with a helpful overview of the literature on preregistration in epidemiology. Thanks to our anonymous reviewers, as well as all of CoAStaL, for useful feedback on the above discussion. Anders S{\o}gaard received financial support from Innovation Fund Denmark, Google Focused Research Awards, and the Novo Nordisk Foundation.

\bibliographystyle{acl_natbib}
\bibliography{References,anthology}

\begin{thebibliography}{52}
\expandafter\ifx\csname natexlab\endcsname\relax\def\natexlab#1{#1}\fi

\bibitem[{Andrade(2021)}]{chittaranjan21hark}
Chittaranjan Andrade. 2021.
\newblock \href {https://doi.org/10.4088/JCP.20f13804} {Harking,
  cherry-picking, p-hacking, fishing expeditions, and data dredging and mining
  as questionable research practices}.
\newblock \emph{The Journal of clinical psychiatry}, 82.

\bibitem[{Baker(2016)}]{baker2016isthere}
Monya Baker. 2016.
\newblock Is there a reproducibility crisis?
\newblock \emph{Nature}, 533:452--454.

\bibitem[{Bakker et~al.(2020)Bakker, Veldkamp, Assen, Crompvoets, Ong, Nosek,
  Soderberg, Mellor, and Wicherts}]{bakker2020}
Marjan Bakker, Coosje Veldkamp, Marcel Assen, Elise Crompvoets, How~Hwee Ong,
  Brian Nosek, Courtney Soderberg, David Mellor, and Jelte Wicherts. 2020.
\newblock \href {https://doi.org/10.1371/journal.pbio.3000937} {Ensuring the
  quality and specificity of preregistrations}.
\newblock \emph{PLOS Biology}, 18:e3000937.

\bibitem[{Barwich(2019)}]{10.3389/fnins.2019.01121}
Ann-Sophie Barwich. 2019.
\newblock \href {https://doi.org/10.3389/fnins.2019.01121} {The value of
  failure in science: The story of grandmother cells in neuroscience}.
\newblock \emph{Frontiers in Neuroscience}, 13.

\bibitem[{Belz et~al.(2021)Belz, Agarwal, Shimorina, and
  Reiter}]{belz-etal-2021-systematic}
Anya Belz, Shubham Agarwal, Anastasia Shimorina, and Ehud Reiter. 2021.
\newblock \href {https://doi.org/10.18653/v1/2021.eacl-main.29} {A systematic
  review of reproducibility research in natural language processing}.
\newblock In \emph{Proceedings of the 16th Conference of the European Chapter
  of the Association for Computational Linguistics: Main Volume}, pages
  381--393, Online. Association for Computational Linguistics.

\bibitem[{Bender et~al.(2021)Bender, Gebru, McMillan-Major, and
  Shmitchell}]{10.1145/3442188.3445922}
Emily~M. Bender, Timnit Gebru, Angelina McMillan-Major, and Shmargaret
  Shmitchell. 2021.
\newblock \href {https://doi.org/10.1145/3442188.3445922} {On the dangers of
  stochastic parrots: {C}an language models be too big?}
\newblock In \emph{Proceedings of the 2021 ACM Conference on Fairness,
  Accountability, and Transparency}, FAccT '21, page 610–623, New York, NY,
  USA. Association for Computing Machinery.

\bibitem[{Bogert et~al.(2021)Bogert, Schecter, and
  Watson}]{bogert2021preregistration}
Eric Bogert, Aaron Schecter, and Rick Watson. 2021.
\newblock \href {https://doi.org/10.17705/1CAIS.04905} {Preregistration of
  information systems research}.
\newblock \emph{Communications of the Association for Information Systems}, 49.

\bibitem[{Bracken(2011)}]{bracken2011}
Michael Bracken. 2011.
\newblock Preregistration of epidemiology protocols: a commentary in support.
\newblock \emph{Epidemiology}, 22:447.

\bibitem[{Bugliarello et~al.(2021)Bugliarello, Cotterell, Okazaki, and
  Elliott}]{bugliarello-etal-2021-multimodal}
Emanuele Bugliarello, Ryan Cotterell, Naoaki Okazaki, and Desmond Elliott.
  2021.
\newblock \href {https://doi.org/10.1162/tacl_a_00408} {Multimodal pretraining
  unmasked: A meta-analysis and a unified framework of vision-and-language
  {BERT}s}.
\newblock \emph{Transactions of the Association for Computational Linguistics},
  9:978--994.

\bibitem[{Card et~al.(2020)Card, Henderson, Khandelwal, Jia, Mahowald, and
  Jurafsky}]{card-etal-2020-little}
Dallas Card, Peter Henderson, Urvashi Khandelwal, Robin Jia, Kyle Mahowald, and
  Dan Jurafsky. 2020.
\newblock \href {https://doi.org/10.18653/v1/2020.emnlp-main.745} {With little
  power comes great responsibility}.
\newblock In \emph{Proceedings of the 2020 Conference on Empirical Methods in
  Natural Language Processing (EMNLP)}, pages 9263--9274, Online. Association
  for Computational Linguistics.

\bibitem[{Chambers(2019)}]{chambers2019whatsnext}
Chris Chambers. 2019.
\newblock \href {https://doi.org/10.1038/d41586-019-02674-6} {What’s next for
  registered reports?}
\newblock \emph{Nature}, 573:187--189.

\bibitem[{Chu and Evans(2021)}]{doi:10.1073/pnas.2021636118}
Johan S.~G. Chu and James~A. Evans. 2021.
\newblock \href {https://doi.org/10.1073/pnas.2021636118} {Slowed canonical
  progress in large fields of science}.
\newblock \emph{Proceedings of the National Academy of Sciences},
  118(41):e2021636118.

\bibitem[{Cohen et~al.(2021)Cohen, Fort, Mieskes, N{\'e}v{\'e}ol, and
  Rogers}]{cohen-etal-2021-reviewing}
Kevin Cohen, Kar{\"e}n Fort, Margot Mieskes, Aur{\'e}lie N{\'e}v{\'e}ol, and
  Anna Rogers. 2021.
\newblock \href {https://doi.org/10.18653/v1/2021.eacl-tutorials.4} {Reviewing
  natural language processing research}.
\newblock In \emph{Proceedings of the 16th Conference of the European Chapter
  of the Association for Computational Linguistics: Tutorial Abstracts}, pages
  14--16, online. Association for Computational Linguistics.

\bibitem[{Cramer(2008)}]{cramer-2008-well}
Irene Cramer. 2008.
\newblock \href {https://aclanthology.org/W08-2206} {How well do semantic
  relatedness measures perform? a meta-study}.
\newblock In \emph{Semantics in Text Processing. {STEP} 2008 Conference
  Proceedings}, pages 59--70. College Publications.

\bibitem[{Devlin et~al.(2019)Devlin, Chang, Lee, and
  Toutanova}]{devlin-etal-2019-bert}
Jacob Devlin, Ming-Wei Chang, Kenton Lee, and Kristina Toutanova. 2019.
\newblock \href {https://doi.org/10.18653/v1/N19-1423} {{BERT}: Pre-training of
  deep bidirectional transformers for language understanding}.
\newblock In \emph{Proceedings of the 2019 Conference of the North {A}merican
  Chapter of the Association for Computational Linguistics: Human Language
  Technologies, Volume 1 (Long and Short Papers)}, pages 4171--4186,
  Minneapolis, Minnesota. Association for Computational Linguistics.

\bibitem[{Dror et~al.(2017)Dror, Baumer, Bogomolov, and
  Reichart}]{dror-etal-2017-replicability}
Rotem Dror, Gili Baumer, Marina Bogomolov, and Roi Reichart. 2017.
\newblock \href {https://doi.org/10.1162/tacl_a_00074} {Replicability analysis
  for natural language processing: Testing significance with multiple
  datasets}.
\newblock \emph{Transactions of the Association for Computational Linguistics},
  5:471--486.

\bibitem[{Field et~al.(2020)Field, Wagenmakers, Kiers, Hoekstra, Ernst, and
  {van Ravenzwaaij}}]{a88772405f594d588336d281cfdf7908}
{Sarahanne M.} Field, E.-J. Wagenmakers, {Henk A. L.} Kiers, Rink Hoekstra,
  {Anja F.} Ernst, and Don {van Ravenzwaaij}. 2020.
\newblock \href {https://doi.org/10.1098/rsos.181351} {The effect of
  preregistration on trust in empirical research findings: Results of a
  registered report}.
\newblock \emph{Royal Society Open Science}, 7(4).

\bibitem[{Gencoglu et~al.(2019)Gencoglu, van Gils, Guldogan, Morikawa, Süzen,
  Gruber, Leinonen, and Huttunen}]{https://doi.org/10.48550/arxiv.1904.07633}
Oguzhan Gencoglu, Mark van Gils, Esin Guldogan, Chamin Morikawa, Mehmet Süzen,
  Mathias Gruber, Jussi Leinonen, and Heikki Huttunen. 2019.
\newblock \href {https://doi.org/10.48550/ARXIV.1904.07633} {Hark side of deep
  learning -- from grad student descent to automated machine learning}.

\bibitem[{Gundersen(2021)}]{Gundersen_2021}
Odd~Erik Gundersen. 2021.
\newblock \href {https://ojs.aaai.org/index.php/aimagazine/article/view/7487}
  {The case against registered reports}.
\newblock \emph{AI Magazine}, 42(1):88--92.

\bibitem[{Gundersen et~al.(2022)Gundersen, Coakley, and
  Kirkpatrick}]{gundersen2022sources}
Odd~Erik Gundersen, Kevin Coakley, and Christine Kirkpatrick. 2022.
\newblock Sources of irreproducibility in machine learning: A review.

\bibitem[{Head et~al.(2015)Head, Holman, Lanfear, Kahn, and
  Jennions}]{10.1371/journal.pbio.1002106}
Megan~L. Head, Luke Holman, Rob Lanfear, Andrew~T. Kahn, and Michael~D.
  Jennions. 2015.
\newblock \href {https://doi.org/10.1371/journal.pbio.1002106} {The extent and
  consequences of p-hacking in science}.
\newblock \emph{PLOS Biology}, 13(3):1--15.

\bibitem[{Hershcovich et~al.(2022)Hershcovich, Frank, Lent, de~Lhoneux, Abdou,
  Brandl, Bugliarello, Cabello~Piqueras, Chalkidis, Cui, Fierro, Margatina,
  Rust, and S{\o}gaard}]{hershcovich-etal-2022-challenges}
Daniel Hershcovich, Stella Frank, Heather Lent, Miryam de~Lhoneux, Mostafa
  Abdou, Stephanie Brandl, Emanuele Bugliarello, Laura Cabello~Piqueras, Ilias
  Chalkidis, Ruixiang Cui, Constanza Fierro, Katerina Margatina, Phillip Rust,
  and Anders S{\o}gaard. 2022.
\newblock \href {https://doi.org/10.18653/v1/2022.acl-long.482} {Challenges and
  strategies in cross-cultural {NLP}}.
\newblock In \emph{Proceedings of the 60th Annual Meeting of the Association
  for Computational Linguistics (Volume 1: Long Papers)}, pages 6997--7013,
  Dublin, Ireland. Association for Computational Linguistics.

\bibitem[{Hovy and Spruit(2016)}]{hovy-spruit-2016-social}
Dirk Hovy and Shannon~L. Spruit. 2016.
\newblock \href {https://doi.org/10.18653/v1/P16-2096} {The social impact of
  natural language processing}.
\newblock In \emph{Proceedings of the 54th Annual Meeting of the Association
  for Computational Linguistics (Volume 2: Short Papers)}, pages 591--598,
  Berlin, Germany. Association for Computational Linguistics.

\bibitem[{Hoyle et~al.(2021)Hoyle, Goel, Hian{-}Cheong, Peskov, Boyd{-}Graber,
  and Resnik}]{DBLP:conf/nips/HoyleGHPBR21}
Alexander~Miserlis Hoyle, Pranav Goel, Andrew Hian{-}Cheong, Denis Peskov,
  Jordan~L. Boyd{-}Graber, and Philip Resnik. 2021.
\newblock \href
  {https://proceedings.neurips.cc/paper/2021/hash/0f83556a305d789b1d71815e8ea4f4b0-Abstract.html}
  {Is automated topic model evaluation broken? the incoherence of coherence}.
\newblock In \emph{Advances in Neural Information Processing Systems 34: Annual
  Conference on Neural Information Processing Systems 2021, NeurIPS 2021,
  December 6-14, 2021, virtual}, pages 2018--2033.

\bibitem[{Ioannidis(2005)}]{10.1371/journal.pmed.0020124}
John P.~A. Ioannidis. 2005.
\newblock \href {https://doi.org/10.1371/journal.pmed.0020124} {Why most
  published research findings are false}.
\newblock \emph{PLoS Med}, 2(8):e124.

\bibitem[{Jager and Leek(2014)}]{Jager2014AnEO}
Leah~Ruth Jager and Jeffrey~T. Leek. 2014.
\newblock An estimate of the science-wise false discovery rate and application
  to the top medical literature.
\newblock \emph{Biostatistics}, 15 1:1--12.

\bibitem[{Johnson(1975)}]{johnson1975models}
Martin Johnson. 1975.
\newblock Models of control and control of bias.
\newblock \emph{European Journal of Parapsychology}, 1:36--44.

\bibitem[{Lash and Vandenbroucke(2012)}]{lash2012should}
{Timothy L} Lash and {Jan P} Vandenbroucke. 2012.
\newblock \href {https://doi.org/10.1097/EDE.0b013e318245c05b} {Should
  preregistration of epidemiologic study protocols become compulsory?:
  Reflections and a counterproposal}.
\newblock \emph{Epidemiology}, 23(2):184--8.

\bibitem[{Lipton and
  Steinhardt(2018)}]{https://doi.org/10.48550/arxiv.1807.03341}
Zachary~C. Lipton and Jacob Steinhardt. 2018.
\newblock \href {https://doi.org/10.48550/ARXIV.1807.03341} {Troubling trends
  in machine learning scholarship}.

\bibitem[{Loder et~al.(2010)Loder, Groves, and Macauley}]{loder10registration}
Elizabeth Loder, Trish Groves, and Domhnall Macauley. 2010.
\newblock \href {https://doi.org/10.1136/bmj.c950} {Registration of
  observational studies}.
\newblock \emph{BMJ (Clinical research ed.)}, 340:c950.

\bibitem[{van Miltenburg et~al.(2021)van Miltenburg, van~der Lee, and
  Krahmer}]{van-miltenburg-etal-2021-preregistering}
Emiel van Miltenburg, Chris van~der Lee, and Emiel Krahmer. 2021.
\newblock \href {https://doi.org/10.18653/v1/2021.naacl-main.51}
  {Preregistering {NLP} research}.
\newblock In \emph{Proceedings of the 2021 Conference of the North American
  Chapter of the Association for Computational Linguistics: Human Language
  Technologies}, pages 613--623, Online. Association for Computational
  Linguistics.

\bibitem[{Nosek et~al.(2018)Nosek, Ebersole, DeHaven, and Mellor}]{Nosek2600}
Brian~A. Nosek, Charles~R. Ebersole, Alexander~C. DeHaven, and David~T. Mellor.
  2018.
\newblock \href {https://doi.org/10.1073/pnas.1708274114} {The preregistration
  revolution}.
\newblock \emph{Proceedings of the National Academy of Sciences},
  115(11):2600--2606.

\bibitem[{Parascandola(2010)}]{parascandola2010epistemic}
Mark Parascandola. 2010.
\newblock \href {https://doi.org/10.1093/lpr/mgq005} {{Epistemic risk:
  empirical science and the fear of being wrong}}.
\newblock \emph{Law, Probability and Risk}, 9(3-4):201--214.

\bibitem[{Perignat and Fleming(2022)}]{https://doi.org/10.1002/anie.202208429}
Elaine Perignat and Fraser~F. Fleming. 2022.
\newblock \href {https://doi.org/https://doi.org/10.1002/anie.202208429}
  {Sunk-cost bias and knowing when to terminate a research project}.
\newblock \emph{Angewandte Chemie International Edition}, 61(36):e202208429.

\bibitem[{Pires et~al.(2019)Pires, Schlinger, and
  Garrette}]{pires-etal-2019-multilingual}
Telmo Pires, Eva Schlinger, and Dan Garrette. 2019.
\newblock \href {https://doi.org/10.18653/v1/P19-1493} {How multilingual is
  multilingual {BERT}?}
\newblock In \emph{Proceedings of the 57th Annual Meeting of the Association
  for Computational Linguistics}, pages 4996--5001, Florence, Italy.
  Association for Computational Linguistics.

\bibitem[{Plank et~al.(2014)Plank, Johannsen, and
  S{\o}gaard}]{plank-etal-2014-importance}
Barbara Plank, Anders Johannsen, and Anders S{\o}gaard. 2014.
\newblock \href {https://doi.org/10.3115/v1/D14-1104} {Importance weighting and
  unsupervised domain adaptation of {POS} taggers: a negative result}.
\newblock In \emph{Proceedings of the 2014 Conference on Empirical Methods in
  Natural Language Processing ({EMNLP})}, pages 968--973, Doha, Qatar.
  Association for Computational Linguistics.

\bibitem[{Rama and Kolachina(2012)}]{rama-kolachina-2012-good}
Taraka Rama and Prasanth Kolachina. 2012.
\newblock \href {https://aclanthology.org/C12-2095} {How good are typological
  distances for determining genealogical relationships among languages?}
\newblock In \emph{Proceedings of {COLING} 2012: Posters}, pages 975--984,
  Mumbai, India. The COLING 2012 Organizing Committee.

\bibitem[{Rid and Schmidt(2010)}]{rid_schmidt_2010}
Annette Rid and Harald Schmidt. 2010.
\newblock \href {https://doi.org/10.1111/j.1748-720X.2010.00474.x} {The 2008
  declaration of helsinki — first among equals in research ethics?}
\newblock \emph{Journal of Law, Medicine \& Ethics}, 38(1):143–148.

\bibitem[{Rowbottom(2011)}]{rowbottom2011kuhn}
Darrell Rowbottom. 2011.
\newblock \href {https://doi.org/10.1016/j.shpsa.2010.11.031} {Kuhn vs. popper
  on criticism and dogmatism in science: A resolution at the group level}.
\newblock \emph{Studies In History and Philosophy of Science Part A},
  42:117--124.

\bibitem[{Rubin(2017)}]{rubin2017when}
Mark Rubin. 2017.
\newblock \href {https://doi.org/10.1037/gpr0000128} {When does harking hurt?
  identifying when different types of undisclosed post hoc hypothesizing harm
  scientific progress}.
\newblock \emph{Review of General Psychology}, 21:308--320.

\bibitem[{Rubin(2020)}]{TQMP16-4-376}
Mark Rubin. 2020.
\newblock \href {https://doi.org/10.20982/tqmp.16.4.p376} {Does preregistration
  improve the credibility of research findings?}
\newblock \emph{The Quantitative Methods for Psychology}, 16(4):376--390.

\bibitem[{Schwab and Held(2020)}]{https://doi.org/10.1111/1740-9713.01369}
Simon Schwab and Leonhard Held. 2020.
\newblock \href {https://doi.org/https://doi.org/10.1111/1740-9713.01369}
  {Different worlds confirmatory versus exploratory research}.
\newblock \emph{Significance}, 17(2):8--9.

\bibitem[{Schwerdtner et~al.(2020)Schwerdtner, Greßner, Kapoor, Assion, Sass,
  Günther, Hüger, and Schlicht}]{https://doi.org/10.48550/arxiv.2011.04328}
Paul Schwerdtner, Florens Greßner, Nikhil Kapoor, Felix Assion, René Sass,
  Wiebke Günther, Fabian Hüger, and Peter Schlicht. 2020.
\newblock \href {https://doi.org/10.48550/ARXIV.2011.04328} {Risk assessment
  for machine learning models}.

\bibitem[{S{\o}gaard(2013)}]{sogaard-2013-estimating}
Anders S{\o}gaard. 2013.
\newblock \href {https://aclanthology.org/N13-1068} {Estimating effect size
  across datasets}.
\newblock In \emph{Proceedings of the 2013 Conference of the North {A}merican
  Chapter of the Association for Computational Linguistics: Human Language
  Technologies}, pages 607--611, Atlanta, Georgia. Association for
  Computational Linguistics.

\bibitem[{S{\o}gaard et~al.(2014)S{\o}gaard, Johannsen, Plank, Hovy, and
  Mart{\'\i}nez~Alonso}]{sogaard-etal-2014-whats}
Anders S{\o}gaard, Anders Johannsen, Barbara Plank, Dirk Hovy, and Hector
  Mart{\'\i}nez~Alonso. 2014.
\newblock \href {https://doi.org/10.3115/v1/W14-1601} {What{'}s in a p-value in
  {NLP}?}
\newblock In \emph{Proceedings of the Eighteenth Conference on Computational
  Natural Language Learning}, pages 1--10, Ann Arbor, Michigan. Association for
  Computational Linguistics.

\bibitem[{S{\o}rensen and Rothman(2010)}]{ae2168802dc011df9806000ea68e967b}
{Henrik Toft} S{\o}rensen and {Kenneth J} Rothman. 2010.
\newblock The prognosis for research.
\newblock \emph{B M J}, 340:c703.

\bibitem[{Strømland(2019)}]{STROMLAND2019102143}
Eirik Strømland. 2019.
\newblock \href {https://doi.org/https://doi.org/10.1016/j.joep.2019.01.006}
  {Preregistration and reproducibility}.
\newblock \emph{Journal of Economic Psychology}, 75:102143.
\newblock Replications in Economic Psychology and Behavioral Economics.

\bibitem[{Ulmer et~al.(2022)Ulmer, Bassignana, M{\"u}ller-Eberstein, Varab,
  Zhang, Hardmeier, and Plank}]{ulmer2022experimental}
Dennis Ulmer, Elisa Bassignana, Max M{\"u}ller-Eberstein, Daniel Varab, Mike
  Zhang, Christian Hardmeier, and Barbara Plank. 2022.
\newblock \href {https://arxiv.org/abs/2204.06251} {Experimental standards for
  deep learning research: A natural language processing perspective}.
\newblock In \emph{ML Evaluation Standards Workshop at ICLR 2022}.

\bibitem[{Veer and Giner-Sorolla(2016)}]{veer2016preregistration}
Anna Veer and Roger Giner-Sorolla. 2016.
\newblock \href {https://doi.org/10.1016/j.jesp.2016.03.004} {Pre-registration
  in social psychology—a discussion and suggested template}.
\newblock \emph{Journal of Experimental Social Psychology}, 67.

\bibitem[{Wiseman et~al.(2019)Wiseman, Watt, and
  Kornbrot}]{wiseman2019registered}
Richard Wiseman, Caroline Watt, and Diana Kornbrot. 2019.
\newblock \href {https://doi.org/10.7717/peerj.6232} {Registered reports: An
  early example and analysis}.
\newblock \emph{PeerJ}, 7:e6232.

\bibitem[{{World Medical Association}(2013)}]{association2013world}
{World Medical Association}. 2013.
\newblock World medical association declaration of helsinki ethical principles
  for medical research involving human subjects.

\bibitem[{Zhao and Bethard(2020)}]{zhao-bethard-2020-berts}
Yiyun Zhao and Steven Bethard. 2020.
\newblock \href {https://doi.org/10.18653/v1/2020.acl-main.429} {How does
  {BERT}{'}s attention change when you fine-tune? an analysis methodology and a
  case study in negation scope}.
\newblock In \emph{Proceedings of the 58th Annual Meeting of the Association
  for Computational Linguistics}, pages 4729--4747, Online. Association for
  Computational Linguistics.

\end{thebibliography}
\end{document}